\documentclass[11pt]{article}

\usepackage[utf8]{inputenc}
\usepackage{lmodern}
\usepackage[T1]{fontenc}
\usepackage[margin=1in]{geometry}
\usepackage{amsmath,amssymb}
\usepackage{booktabs}
\usepackage{graphicx}
\usepackage{hyperref}
\usepackage{xurl}       
\usepackage[style=nejm, citestyle=numeric-comp, sorting=none, backend=biber]{biblatex}
\usepackage[expansion=false]{microtype}
\usepackage{array}
\usepackage{xcolor}
\usepackage{caption}
\usepackage{subcaption}
\usepackage{float}
\usepackage{placeins}
\usepackage{setspace}
\usepackage{parskip}
\usepackage{lineno}

\graphicspath{{figures/}{./}}

\title{\textbf{From Prediction to Self:\\
Developmental Conditions for Agency in Minimal Neural Systems}}

\author{%
  Evan Ye\\[0.4ex]
  \normalsize Independent Researcher\\[0.25ex]
  \normalsize \texttt{ye635498222@gmail.com}%
}
\date{}

\begin{document}
\color{black}
\maketitle

\begin{abstract}
How does a system that merely predicts the world come to distinguish its own causal
influence from everything else? We trace this transition in a minimal 192-dimensional
GRU through a developmental sequence --- 6 experimental stages, 12
falsified alternatives, and cross-signal validation. Starting with no action or
self-representation, we add components one at a time, tracking whether the system
distinguishes self-caused from world-caused changes. The central finding is the \emph{encoding gap}: a system
can perfectly compensate for its own actions in prediction while failing to encode
``I am acting'' as a readable state --- implicit causal use and explicit
self-representation are dissociated capabilities.

The developmental path crosses this gap when four conditions are jointly satisfied:
(1) persistent state that forms stable attractors, (2) a causal action
loop linking the system's output to its input, (3) proprioceptive feedback that makes
implicit causal knowledge explicit, and (4) asynchronous awakening — consolidating
perceptual learning before action learning, which yields the only configuration robust to
hyperparameter choice. We propose \emph{agency gain}
($\mathcal{A} = \text{Err}_{\text{world}} - \text{Err}_{\text{self}}$), the predictive
advantage of knowing one's own action, as a continuous metric that generalizes across
signal types.

A decisive test confirms the causal grounding of the encoding: after the external training
signal is removed, the causal agent retains its self-representation at 94.9\% while a
statistically-matched control collapses to 53.9\%. Self-representation persists only when
causally useful for prediction --- an intrinsic property of the causal loop, not a training
artifact.
\end{abstract}


\section{Introduction}

A system that predicts the weather does not know it is predicting the weather. It maps
inputs to outputs — past observations to future estimates — without any representation
of itself as a distinct causal entity in the process. Adding actions changes this
situation fundamentally: when a system's outputs feed back into its inputs through the
world, the observations it receives are no longer purely external. Some of the changes
it observes are consequences of what it did. The question this paper addresses is:
under what conditions does a predictive system come to distinguish these self-caused
changes from world-caused changes?

This question sits at the intersection of several research programs. Predictive
processing and the free energy principle~\cite{friston2010} propose that organisms
minimize prediction error through perception and action, with efference
copies~\cite{vonholst1950} canceling self-caused sensory changes.
Empowerment~\cite{klyubin2005} measures the channel capacity between actions and future
states. Curiosity-driven exploration~\cite{pathak2017} uses prediction error as
intrinsic reward. Developmental robotics~\cite{oudeyer2007} studies how sensorimotor
competence unfolds over time.

What is missing from these programs is a systematic, empirical account of how self-world
decomposition develops from scratch — what the minimal conditions are, what order they
must appear in, and what plausible-sounding alternatives fail. The theoretical proposals
are rich, but the experimental base is thin: we do not know, in any concrete system,
the precise boundary between ``a system that predicts'' and ``a system that knows it is
the one predicting.''

This paper provides that account. Using a minimal architecture --- a 192-dimensional GRU
with multi-scale dynamics, fewer than 100K parameters --- we conduct a systematic
developmental sequence of experiments. Starting from a system with no action
and no self-representation, we add components one at a time and observe what changes.
Each experiment is motivated by a question that the previous experiment raised, and each
is evaluated against pre-specified scorecards with explicit PASS/FAIL criteria.

The central finding is the \emph{encoding gap}: a system can causally use its own action in
prediction yet fail to encode it as a readable state --- causal use and self-representation are
dissociated capabilities. Self-world decomposition crosses this gap when four conditions are
jointly satisfied, in this order: persistent state, causal action loop, proprioceptive feedback, and
asynchronous awakening. Twelve alternative approaches all fail, for reasons we
characterize precisely. To quantify the decomposition at each stage, we propose agency
gain: the gap in prediction error between a model that knows the system's action and an
otherwise-identical model that does not. This metric is measurable, ablatable,
comparable, and environment-agnostic.

By \emph{self-representation} we mean a state variable in $h_{\text{multi}}$ that is
(i)~\emph{causal} --- its presence measurably improves prediction of self-caused
observations; (ii)~\emph{readable} --- linearly decodable from the persistent state; and
(iii)~\emph{persistent} --- maintained by the system's own dynamics without ongoing
external supervision. Whether such a variable forms, and under what developmental
conditions, is the empirical question this paper addresses.

We make no claims about consciousness, sentience, or subjective experience. We claim
only to have mapped the conditions under which a minimal predictive system learns to
distinguish its own causal influence from the rest of its world.

\section{Method}

\subsection{Experimental Paradigm}

All experiments share a common structure. A world produces a multi-channel signal.
A model receives the signal, maintains persistent internal state, and predicts the
next observation. We intervene — adding an action loop, changing the architecture,
ablating a component — and observe the consequences through quantitative metrics.

Each experiment changes one variable relative to its predecessor. This allows precise
attribution: if metric $X$ changes when component $Y$ is added, and only when component
$Y$ is added, the change is attributable to $Y$.

\subsection{World}

The default environment is a 4-channel sinusoidal signal. Multiple channels are required
so that the system can, in principle, distinguish the channel it affects from those it
does not; a single-channel design would confound self-caused and world-caused variation.
Each channel consists of two frequency components with distinct base frequencies and
amplitudes, plus Gaussian noise ($\sigma = 0.05$). The frequencies are coprime, giving a
long compound period ($10{,}000$ steps) relative to the short windows the probes use, so
short-horizon prediction cannot reduce to memorizing a cycle; the additive noise makes
the signal impossible to predict exactly. When an action loop is present, the action modifies one channel:
$\text{obs}[0] \mathrel{+}= \gamma \cdot a(t)$, where $\gamma = 2.0$.

\subsection{Model}

The core model is a GRU (Gated Recurrent Unit) with 192 hidden dimensions, augmented
with multi-scale exponential moving average (EMA). We adopt the GRU as the core
dynamical unit motivated by a fundamental requirement of our investigation: any system
capable of developing self-world decomposition must maintain persistent internal state
that evolves continuously along the temporal axis. Unlike feedforward architectures
that process segmented inputs independently, the GRU maintains an ongoing hidden state
$h_t$ that never resets, naturally instantiating the closed sensorimotor loop through
which the system's outputs feed back into its inputs. GRU is chosen over LSTM for its
simpler gating structure (25\% fewer parameters), aligning with our minimality
principle --- the fewest architectural degrees of freedom that isolate the phenomenon
of interest, avoiding confounds from unused capacity. The hidden dimension of 192 (48 per multi-scale EMA group over 4 timescales) is a design default rather than a
tuned or minimized value. It is comfortably adequate for the low-complexity signals
studied here (effective attractor dimensionality $\approx$5,
Section~\ref{sec:perception}); in principle reservoir capacity should scale with signal
complexity, so higher-dimensional or longer-timescale inputs would require more
(Section~\ref{sec:limitations}). A systematic size sweep is left for future work. In the final configuration, the GRU weights receive zero gradient through
the detach mechanism (Section~\ref{sec:training}), functioning as a fixed reservoir;
only the readout heads (\texttt{pred\_A}, \texttt{pred\_B}, \texttt{W\_action}) are
trained. This design choice is discussed further in
Section~\ref{sec:discussion-reservoir}.

Multi-scale EMA is computed as:
\begin{equation}
  h_{\text{multi}}(t) = (\mathbf{1} - \boldsymbol{\alpha}) \odot h_{\text{multi}}(t-1)
    + \boldsymbol{\alpha} \odot \text{GRU}\!\left(x(t),\, h_{\text{gru}}(t-1)\right)
\end{equation}
where $\boldsymbol{\alpha} \in \mathbb{R}^{192}$ is a per-dimension gain vector
partitioned into four blocks of 48 dimensions; each block takes a constant scalar drawn
from a log-spaced grid spanning 0.02 (slow, long memory) to 0.80 (fast, immediate
response). The operator $\odot$ denotes element-wise (Hadamard) multiplication, so
different dimensions integrate over different timescales in parallel. Multiple timescales provide
the system with natural temporal hierarchy, echoing the multi-timescale organization
of biological memory systems --- fast dynamics track immediate change, slow dynamics
maintain contextual continuity. The state $h_{\text{multi}}$
is never reset --- it runs continuously across all training steps, accumulating a
persistent representation of the system's entire history.

A prediction head maps $h_{\text{multi}}$ to the predicted next observation. When dual
heads are present (Section~\ref{sec:measurement}), \texttt{pred\_A} receives both
$h_{\text{multi}}$ and the action value, while \texttt{pred\_B} receives only
$h_{\text{multi}}$.

\begin{figure}[htbp]
  \centering
  \includegraphics[width=0.9\linewidth]{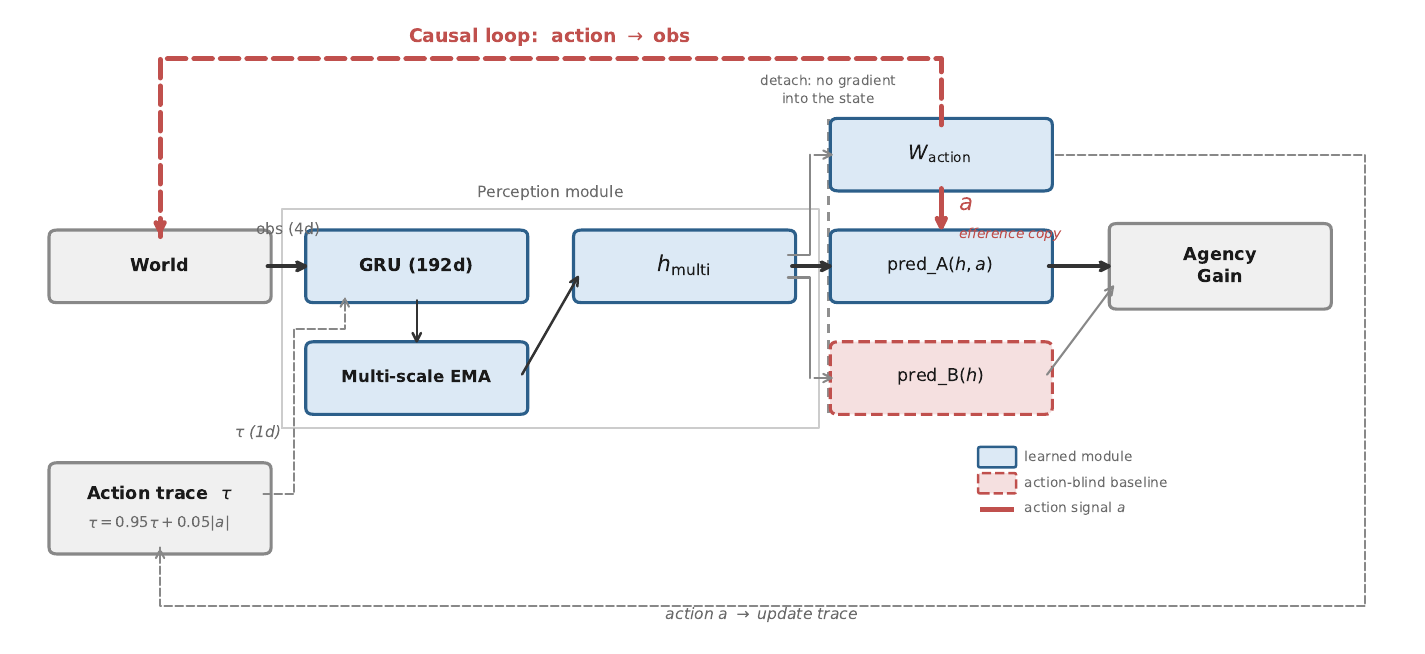}
  \caption{System architecture. The main trunk --- World $\to$ GRU $\to$
    $h_\text{multi}$ $\to$ pred\_A $\to$ Agency Gain --- runs left to right; the
    multi-scale EMA, $W_\text{action}$, and the action trace $\tau$ are branches. The
    action-aware predictor pred\_A$(h,a)$ receives the action $a$ as an explicit
    efference-copy input (bold red), while the action-blind baseline pred\_B$(h)$ does
    not; the \emph{same} $a$ closes the causal loop back onto the world (bold red
    dashed), so the observation carries self-caused structure. Agency Gain
    $\mathcal{A} = \text{Err}_\text{world} - \text{Err}_\text{self}$ is the divergence of
    the two predictors. The detach boundary marks where prediction-head gradients do not
    flow back into the state. \textit{Not all components are active in every experiment;
    see Table~\ref{tab:config} for the per-stage configuration.}}
  \label{fig:architecture}
\end{figure}

The system architecture is illustrated in Figure~\ref{fig:architecture}.

\FloatBarrier
\subsection{Causal vs.\ Control Design}

Experiments that test causal attribution use a matched-control design. The Causal group
has action $a(t) = f(h_{\text{multi}})$ — the action is a function of the system's
internal state, creating a genuine causal loop. The Control group replaces this with
AR(1) noise matched in mean, variance, and autocorrelation to the Causal group's
actions. This preserves the statistical properties of the signal while breaking the
causal link between the system's state and its action.

\subsection{Agency Gain}

To quantify self-world decomposition, we define agency gain as the predictive advantage
of knowing one's own action:
\begin{equation}
  \mathcal{A}(t) =
    \underbrace{\bigl\|o(t{+}1) - \hat{o}_{\text{world}}(t{+}1)\bigr\|^2}%
               _{\text{Err}_{\text{world}}}
    -
    \underbrace{\bigl\|o(t{+}1) - \hat{o}_{\text{self}}(t{+}1)\bigr\|^2}%
               _{\text{Err}_{\text{self}}}
\end{equation}
where $\hat{o}_{\text{self}}$ is the prediction of a model that knows the action, and
$\hat{o}_{\text{world}}$ is the prediction of an otherwise-identical model that does
not. When $\mathcal{A} > 0$, the action carries causal information that improves
prediction. The prediction gap is defined as
$(\text{Err}_{\text{world}} - \text{Err}_{\text{self}}) / \text{Err}_{\text{world}}$.

The spike test verifies causality by replacing \texttt{pred\_A}'s action input with zero
(while the action continues to affect the observation normally) and measuring the increase
in $\text{Err}_{\text{self}}$ over a measurement window (3{,}000 steps):
\begin{equation}
  \text{spike} =
    \frac{\text{Err}_{\text{self}}^{\text{action disconnected}}}%
         {\text{Err}_{\text{self}}^{\text{normal}}}
\end{equation}
A spike significantly greater than 1.0 confirms that the prediction advantage depends
on the action's causal effect, not on statistical artifacts.

We use \emph{agency} to refer to a system's capacity to causally influence its
observations through its actions. A system has agency with respect to an observation
channel when its actions produce a measurable causal effect on that channel ---
operationally, when both agency gain $\mathcal{A} > 0$ and the spike test confirms
causal dependence (spike significantly greater than~1.0). Agency gain is the metric;
the spike test is the causal verification; agency is the property they jointly
diagnose.

\subsection{Training Protocol}
\label{sec:training}

Training follows a predict-then-update protocol: at each step, the model first predicts
the current observation from its old state, then computes the loss, then updates weights
via backpropagation, and finally updates $h_{\text{multi}}$ with the current observation
(under no gradient). This prevents information leakage — the model cannot ``see'' the
observation before predicting it.

State updates are performed without gradient tracking to prevent backpropagation through
time across the action pathway, ensuring that action-learning gradients do not corrupt
perceptual representations.

\section{Results: The Developmental Path}

\begin{figure}[htbp]
  \centering
  \includegraphics[width=0.9\linewidth]{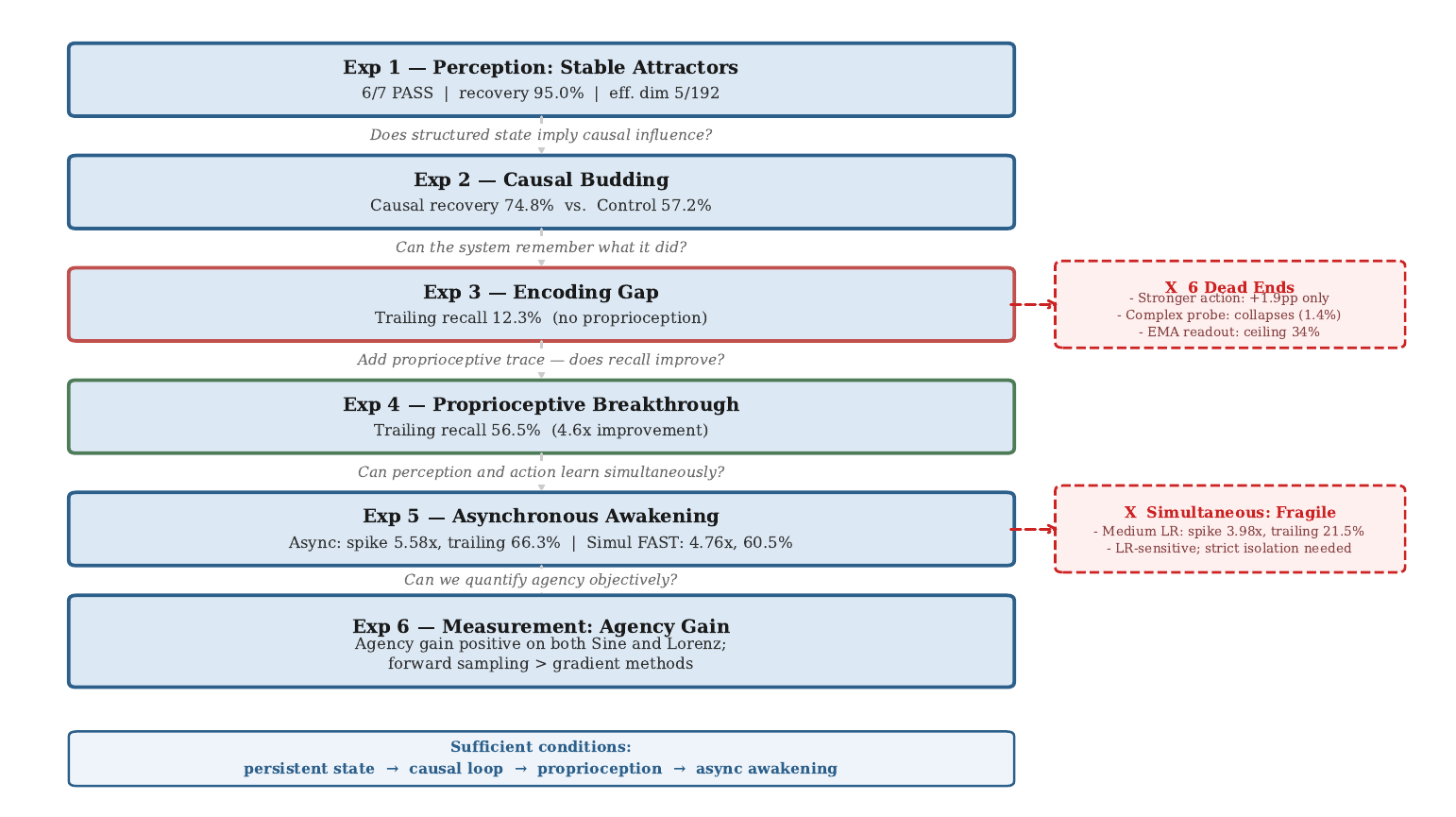}
  \caption{Developmental chain of the six experiments. Red-bordered box (Exp.\ 3) marks
    the Encoding Gap bottleneck; green-bordered box (Exp.\ 4) marks the Proprioceptive
    Breakthrough. Right-hand dashed boxes show dead-end variants. The bottom bar states
    the four sufficient conditions.
    Self-maintenance (Exp.\ 4b): Causal 94.9\% vs.\ Control 53.9\%.}
  \label{fig:chain}
\end{figure}

Figure~\ref{fig:chain} shows the complete developmental chain of the six experiments.
Each box corresponds to one stage; the chain proceeds left-to-right. Red-bordered box
marks the Encoding Gap (Exp.\ 3); green-bordered box marks the Proprioceptive
Breakthrough (Exp.\ 4).

Two configuration choices recur across the sequence and are summarized in
Table~\ref{tab:config}. The GRU is held frozen (a fixed reservoir) wherever the
aim is to \emph{measure} a property of $h_{\text{multi}}$, so that the measured
information reflects the fixed dynamics rather than a training artifact
(Section~\ref{sec:discussion-reservoir}); it is \emph{trained}, via an auxiliary loss on
the live state during Phase~2, only where the aim is to actively pressure the state
toward an explicit self-encoding --- and the encoding gap (Section~\ref{sec:encoding})
persists even under that pressure. Actions are externally random during exploration and
self-generated once an action policy is engaged.

\begin{table}[htbp]
\centering
\footnotesize
\caption{Configuration of each developmental stage. ``Policy'' denotes a self-generated
action $a = W_{\text{action}}(h)$; ``aux'' an auxiliary classification head applied to
the live state; a checkmark marks the proprioceptive trace input. ``GRU'' states whether
the recurrent weights receive gradient. ``Schedule'' gives the phase structure ---
training phases (P1, P2, \ldots; step counts at full scale) for
Sections~\ref{sec:self-maintenance}--\ref{sec:measurement}, or the run/test protocol for
the perception and causal-budding experiments. The action policy $W_{\text{action}}$ is
frozen (random) through Sections~\ref{sec:perception}--\ref{sec:proprioception} and is
first trained in Section~\ref{sec:async} (asynchronous awakening).}
\label{tab:config}
\setlength{\tabcolsep}{4pt}%
\begin{tabular}{@{}llccll@{}}
\toprule
Stage & Heads & Trace & Actions & GRU & Schedule \\
\midrule
3.1 Perception         & single                     & ---        & none                & frozen           & single run, 50K \\
3.2 Causal budding     & single                     & ---        & policy / AR(1)      & frozen           & warmup $\to$ disconnect \\
3.2.1 Self-maintenance & single + aux               & ---        & policy (sparse)     & \textbf{trained} & P1 100K $\to$ P2 80K $\to$ P3 5K \\
3.3 Encoding gap       & single + aux               & ---        & random              & \textbf{trained} & P1 100K $\to$ P2 80K \\
3.4 Proprioception     & single + aux               & \checkmark & random              & \textbf{trained} & P1 100K $\to$ P2 80K \\
3.5 Async awakening    & dual + $W_{\text{action}}$  & \checkmark & random $\to$ policy & \textbf{trained} & P1 100K $\to$ P2a 60K $\to$ P2b 60K \\
3.6 Measurement        & dual + $W_{\text{action}}$  & optional   & random $\to$ policy & frozen           & P1 100K $\to$ P2a 60K $\to$ P2b 60K \\
\bottomrule
\end{tabular}
\end{table}

\FloatBarrier
\subsection{Perception: Stable Attractors}
\label{sec:perception}

We begin with the simplest possible system: a continuously-running GRU predicting a
sinusoidal signal. No action, no self/world distinction — just prediction. The question
is whether persistent prediction alone produces stable internal structure.

\begin{figure}[htbp]
  \centering
  \includegraphics[width=0.9\linewidth]{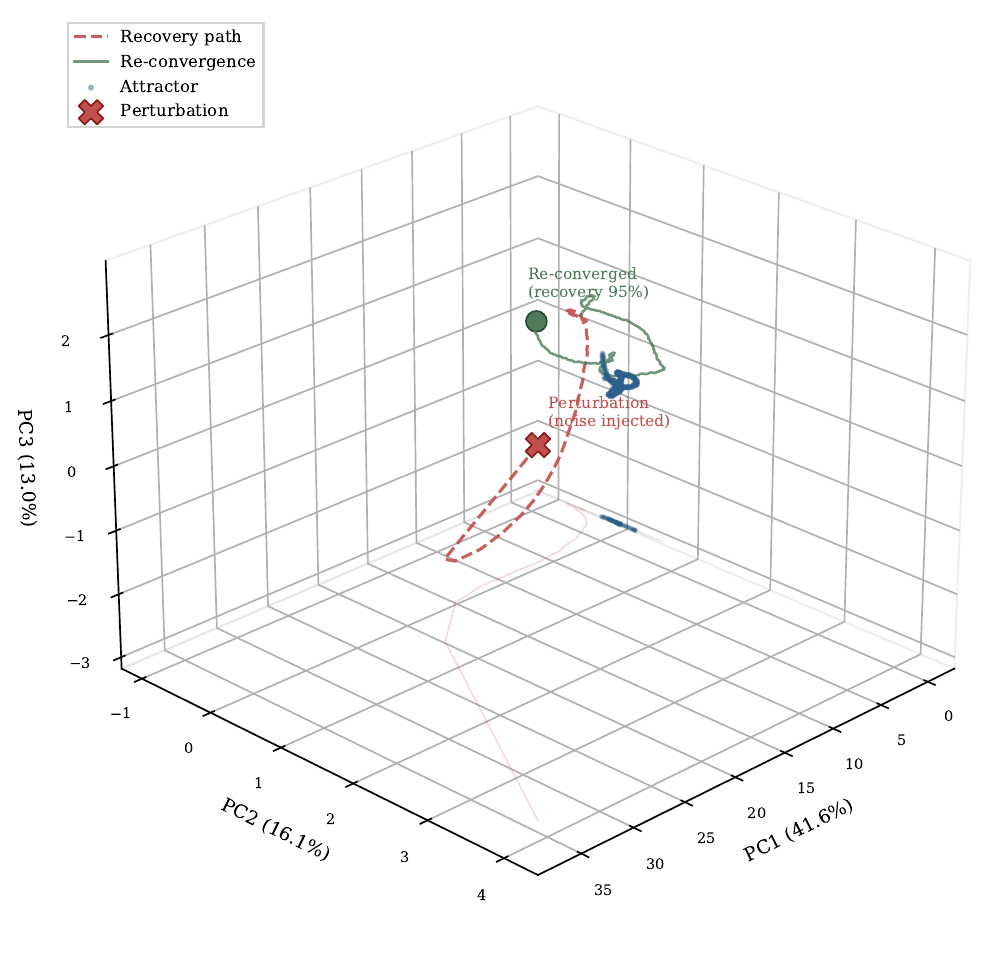}
  \caption{PCA projection (3 components, 70.8\% variance) of $h_\text{multi}$ across
    30\,000 training steps on the sinusoidal signal. Blue cloud: stable attractor region.
    Red dashed path: recovery trajectory after noise perturbation at step 15\,000. Green
    path: re-convergence. Recovery rate: 95.0\%.}
  \label{fig:attractor}
\end{figure}

\textbf{Result.} The system forms stable low-dimensional attractors (Figure~\ref{fig:attractor}).
PCA analysis reveals that 95\% of the variance in $h_{\text{multi}}$ is concentrated in
5 dimensions out of 192 (effective dimensionality $< 3\%$). The multi-scale EMA
produces hierarchical temporal structure: power spectrum analysis confirms that the four
timescale groups specialize in different frequency bands. Perturbation recovery reaches
95.0\% — after injecting noise into $h_{\text{multi}}$, the system returns to its
attractor within hundreds of steps. The scorecard passes 6 of 7 tests (the exception being residual autocorrelation,
indicating the GRU has not fully extracted all predictable structure from the signal).

\textbf{Significance.} Persistent prediction with multi-scale dynamics is sufficient to
form stable, low-dimensional, perturbation-resistant internal structure. This structure
is the foundation for everything that follows.

\textbf{Question raised.} The system has stable internal structure but no action. What
happens when its behavior can change the world?

\FloatBarrier
\subsection{Causal Budding: Implicit Self-World Decomposition}
\label{sec:causal}

We add a causal action loop: the system generates an action $a(t) = f(h_{\text{multi}})$
through a linear projection, and this action modifies one observation channel:
$\text{obs}[0] \mathrel{+}= \gamma \cdot a(t)$, with $\gamma = 2.0$. Channels 1–3 are
unaffected. The system still has a single prediction head. The question is whether the
system implicitly learns which observation changes are self-caused.

\begin{figure}[htbp]
  \centering
  \includegraphics[width=0.9\linewidth]{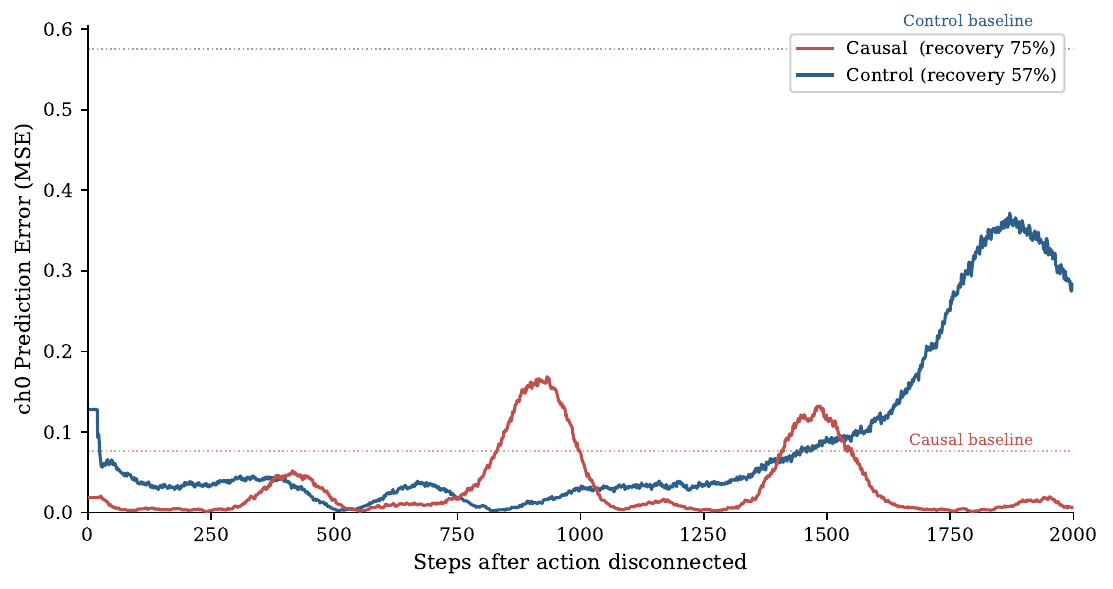}
  \caption{Channel-0 prediction error (MSE, 20-step rolling average) after action
    disconnection. The causal model (red) recovers to near-baseline whereas the control
    (blue, AR-process action) drifts. Recovery metric: causal 74.8\% vs.\ control 57.2\%.}
  \label{fig:spike}
\end{figure}

\textbf{Result.} Disconnecting the action and comparing the Causal and Control groups
over a long-disconnect test (2,000 steps with the action removed) reveals an asymmetry:
the Causal group recovers to 74.8\% of baseline prediction accuracy, whereas the Control
group recovers only 57.2\% (Figure~\ref{fig:spike}). Because both groups experience the
same distributional shift (the Control group's action is AR(1) noise with matched statistics), the difference is attributable to causal structure rather than
distributional surprise. This is consistent with self/world decomposition, but the
decisive evidence comes from the self-maintenance test below: only the causally-grounded
encoding survives removal of the training signal.

\textbf{Significance.} A single-head predictive system with a causal action loop
implicitly learns self-world decomposition. This decomposition is real (verified by
Causal vs.\ Control), channel-specific, and emerges without any architectural prior.
We use \emph{decomposition} here in the operational sense of Causal-vs-Control divergence
at the behavioral level; the stronger claim of \emph{representational} decomposition (an explicitly readable ``I am acting'' state in $h_{\text{multi}}$) is established
below (Sections~\ref{sec:encoding}--\ref{sec:proprioception}).

\textbf{Question raised.} The system ``knows'' implicitly which changes it causes. But
can a probe read ``I am acting'' from $h_{\text{multi}}$?

\subsubsection{Self-Maintenance: Selection Pressure on Representations}
\label{sec:self-maintenance}

The result above shows that the Causal group learns implicit self-world decomposition
--- $h_{\text{multi}}$ encodes information about the system's own action effect. But
does this encoding depend on the external training signal, or is it self-sustaining
once formed?

To test this, we ran an ablation using the original architecture (3-channel signal, no
trace input, a binary auxiliary loss that explicitly forces the GRU to encode whether
it is currently acting --- the burst-active state; the burst gate alternates between
acting and quiet periods, detailed in
Section~\ref{sec:encoding}).\footnote{This experiment uses the same burst-gate mechanism
but in a different setting from the failed Hypothesis~12 in
Section~\ref{sec:falsified} (2-D grid, counterfactual directional prediction, no valid
action channel during stillness); the two results are not contradictory.}
Phase~2 pushes the Causal group's classification accuracy to 95.3\% and the Control
group's to 91.9\%. We then remove the auxiliary loss and run 5{,}000 additional steps
with all parameters frozen.

\begin{center}
\textbf{The Causal group retains 94.9\% accuracy (near-lossless), while the Control
group collapses to 53.9\% --- near chance.}
\end{center}

This asymmetry is decisive. In both groups, the system was pushed to encode the same
state variable during training, and both achieved comparable near-perfect accuracy.
What differs after the training signal is removed is whether the encoding continues
to serve any function. In the Causal group, the encoding is entangled with the ongoing
prediction task through the causal loop --- the system's action affects $\text{obs}[0]$,
and predicting $\text{obs}[0]$ requires representing the action state. The GRU's
learned dynamics maintain the encoding without external pressure because it remains
predictively useful. In the Control group, the action is statistically matched but
causally decoupled; the encoding has no predictive utility, and it decays the moment
external supervision is removed. Notably, in this experiment the action affects the
observation with a two-step delay ($\tau = 2$): predicting $\text{obs}[0]$ requires the
dynamics to carry recent action history across the gap, so the encoding's predictive
usefulness (and hence its persistence) is not an artifact of immediate causation.
The variable retained here is the \emph{currently-acting} (burst-active) state, which
stays anchored in the present observation; the harder \emph{recently-acted} (trailing)
state, whose anchor vanishes once the action ceases, does not self-sustain as robustly
(Section~\ref{sec:falsified}, H6: balanced accuracy $80\% \to 70\%$) --- consistent with
the encoding gap.

This finding suggests a \emph{selection pressure} on self-representations: they are
sustained by the system when, and only when, they are causally useful for
prediction. Section~\ref{sec:discussion-sustained} discusses the implications of this
asymmetry for theories of self-representation stability.

\FloatBarrier
\subsection{The Encoding Gap}
\label{sec:encoding}

Section~\ref{sec:self-maintenance} showed that $h_{\text{multi}}$ can be trained to
encode the \emph{current} acting state with high accuracy (Causal $94.9\%$ retention
after aux loss removal). Here we test a stricter question: after action ceases, does
$h_{\text{multi}}$ retain a \emph{recently acted} signal in a readable form?

We introduce a 3-class labeling scheme aligned to the burst gate: ACTIVE (currently
acting), TRAILING (within the 50-step window after action ceases), and QUIET (long
inactive period). An auxiliary 3-class classification head propagates gradient into
the GRU during Phase~2 to explicitly pressure it toward encoding these categories,
and a separate detached linear probe measures how well the TRAILING class can be
read from $h_{\text{multi}}$.

\textbf{Result.} Even under this direct gradient pressure, \emph{trailing recall} ---
the probe's ability to detect ``recently acted'' during the 50-step window after burst
gate deactivation --- reaches only 12.3\%.
The system can perfectly compensate for its own actions in prediction, but $h_{\text{multi}}$
does not retain ``I was acting'' in a readable form once action stops.

Five successive experiments attempt to break through this ceiling (Table~\ref{tab:gap}):

\begin{table}[H]
\centering
\small
\caption{Failed attempts to bridge the encoding gap.}
\label{tab:gap}
\begin{tabular}{l >{\raggedright\arraybackslash}p{3.5cm} >{\raggedright\arraybackslash}p{5.5cm}}
\toprule
Attempt & Hypothesis & Result \\
\midrule
Stronger action ($\gamma{\times}1.5$) & Larger effect $\to$ clearer encoding & $+1.9$pp \\
Passive EMA readout & Slow EMA retains trace & Ceiling 34\% \\
GRU probe (complex) & Nonlinear probe extracts signal & 1.4\% \\
Character-level CE & Train on rare ``i'' tokens & 0.0\% \\
Stronger action ($\gamma{\times}2.0$) & Even larger effect & Marginal \\
\bottomrule
\end{tabular}
\par\smallskip
\begin{minipage}{0.92\textwidth}
\footnotesize\textit{Row-by-row takeaways:}
(1)~Implicit compensation scales with action strength but hits the same ceiling regardless of scale.
(2)~Exponential decay erases action history; slow EMA cannot retain the trace.
(3)~Signal is not weak --- it is absent: a nonlinear probe finds nothing to extract.
(4)~Gradient dominated by 99\% majority class; rare tokens carry no useful gradient.
(5)~Same ceiling, different scale: stronger action shifts magnitude but not the ceiling.
\end{minipage}
\end{table}

As a further control, a linear probe trained to read the binary \emph{is-acting} state
directly from $h_{\text{multi}}$ (without auxiliary-head pressure), evaluated across all
periods, reaches only $70\%$. This residual is carried by the current observation during
active periods (when the action is still present in the input and thus trivially detectable) rather than by an explicit representation in $h_{\text{multi}}$. Once the
action ceases (the trailing window), readability collapses to the $12.3\%$ above:
implicit leakage through the current input is not explicit self-representation.

\textbf{Key finding.} Implicit causal knowledge is not the same as explicit
self-representation. Prediction accuracy and self-representation are dissociated.

\textbf{Question raised.} What is missing? What would make the implicit explicit?

\subsection{Proprioceptive Breakthrough}
\label{sec:proprioception}

We add a single architectural change: an action trace as proprioceptive input to the GRU.
\begin{equation}
  \tau(t) = \beta \cdot \tau(t-1) + (1 - \beta) \cdot |a(t)|, \quad \beta = 0.95
\end{equation}
The GRU input changes from 4 dimensions (obs only) to 5 dimensions (obs + action trace).
No other change.

\textbf{Result.} Trailing recall jumps from 12.3\% to 56.5\%. A single additional input
dimension breaks through a ceiling that six prior experiments could not move.

\textbf{Symbol grounding.} This explicit representation enables grounding the
first-person pronoun ``i'' to causal dynamics. During trailing periods, the system is
presented with character sequences such as ``i moved''; during quiet periods, it sees
``the world changed.'' A linear probe achieves balanced accuracy of 80.1\% (chance
$= 50\%$), compared to 64.4\% for the Control group ($+15.7$pp gap). Generalization to
unseen sentences (``i jumped,'' ``i stopped'') reaches 83.8\%, demonstrating that the
probe learned the mapping from trailing-state dynamics to ``i,'' not from specific word
patterns~\cite{harnad1990}. We report this as a first demonstration that the readable
self-state can anchor a symbol; a compositional treatment of grounding is developed in
future work.

\textbf{Significance.} The encoding gap requires a new information channel: proprioceptive
feedback that directly writes action history into the system's state. Proprioception
transforms implicit causal knowledge (distributed in the weights) into explicit
self-representation (readable from $h_{\text{multi}}$).

\textbf{Question raised.} The system now has explicit self-representation. Can it learn
to actively control its actions?

\subsection{Asynchronous Awakening}
\label{sec:async}

We unfreeze $W_{\text{action}}$ — the linear projection from $h_{\text{multi}}$ to
action — and attempt to train it alongside the perceptual system.
To select structured actions, we introduce the dual-head architecture here:
\texttt{pred\_A} receives $h_{\text{multi}}$ and the action value;
\texttt{pred\_B} receives only $h_{\text{multi}}$.
At each step the action that maximises the disagreement between the two heads is chosen
(forward-sampled disagreement maximisation).
Both the asynchronous and simultaneous conditions use this same dual-head setup;
the only difference between them is the training schedule, not the architecture.

\textbf{Result: simultaneous learning is unstable.} Three experiments with different
action learning rates show inconsistent results (Table~\ref{tab:simul}):

\begin{table}[H]
\centering
\small
\caption{Simultaneous learning results by learning rate.}
\label{tab:simul}
\begin{tabular}{@{}lrrl@{}}
\toprule
LR & Spike & Trailing & Pass? \\
\midrule
Fast ($10^{-3}$) & $4.76\times$ & 60.5\% & Yes \\
Medium ($5{\times}10^{-4}$) & $3.98\times$ & 21.5\% & \textbf{No} \\
Slow ($10^{-4}$) & $2.52\times$ & 40.3\% & Yes (marginal) \\
\bottomrule
\end{tabular}
\end{table}

No simultaneous configuration achieves the best result on both metrics.

\textbf{Solution: temporal separation.} Training is divided into three phases:
\begin{itemize}\setlength{\itemsep}{2pt}
  \item \textbf{Phase 1} (100K steps): Random actions. $W_{\text{action}}$ frozen. LR $= 10^{-3}$.
  \item \textbf{Phase 2a} (60K steps): $W_{\text{action}}$ still frozen. LR drops to $10^{-4}$. Perception consolidates.
  \item \textbf{Phase 2b} (60K steps): $W_{\text{action}}$ unfrozen. Action policy trains on a stable perceptual foundation.
\end{itemize}

\textbf{Result: asynchronous learning produces the best result.} Spike ratio $5.58\times$
(highest of all configurations), trailing recall $66.3\%$ (highest of all configurations).

\textbf{Significance.} Temporal separation of perception and action learning produces
the most robust results. The developmental sequence — perception first, then action —
is the reliable path~\cite{rochat2001}.

\textbf{Question raised.} How do we quantify how much knowing one's own action
actually helps?

\subsection{Measurement: Agency Gain}
\label{sec:measurement}

The dual-head architecture, already introduced in Section~\ref{sec:async} for action
selection, now serves a measurement purpose: the prediction gap between
\texttt{pred\_A} and \texttt{pred\_B} directly quantifies agency gain.
\texttt{pred\_A} receives $h_{\text{multi}}$ and the action; \texttt{pred\_B} receives
only $h_{\text{multi}}$. Both are trained to predict the same target. Actions continue to
be selected via forward-sampled disagreement maximisation: from four candidates
(action base, base $\pm$ perturbation, zero), the system executes whichever produces the
largest difference between \texttt{pred\_A} and \texttt{pred\_B} predictions.

\begin{figure}[htbp]
  \centering
  \includegraphics[width=0.9\linewidth]{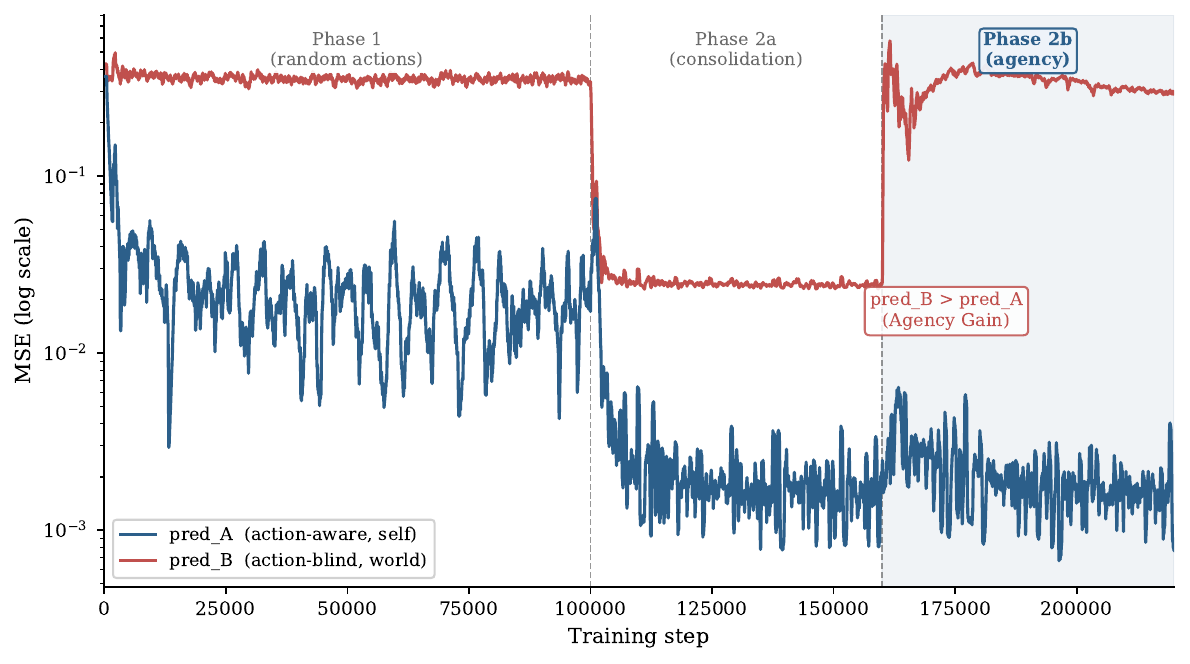}
  \caption{Three-phase training curves on the Lorenz signal (log MSE), under the
    forward-sampled action-selection strategy (the gradient-trained alternatives are
    compared in Table~\ref{tab:strategy}).
    \texttt{pred\_A} (blue, action-aware) remains below \texttt{pred\_B}
    (red, action-blind) across all three phases.
    The Phase~1 gap reflects mechanical compensation of known random perturbations
    (no causal dependence); the Phase~2b gap accompanies structured action selection
    (genuine causal dependence). Agency gain is measurable and positive throughout.
    Spike ratios quantifying the causal dependence are reported in
    Sections~\ref{sec:measurement} and~\ref{sec:discussion-thermometer}.}
  \label{fig:agency}
\end{figure}

\textbf{Action strategy comparison.}
The two gradient-based action-training baselines reported in the preprint (arXiv~v1)
contained implementation errors --- the perception loss was entangled with the action
objective, and no gradient reached $W_{\text{action}}$. The corrected results are reported
here; the forward-sampled strategy was unaffected by the correction. Corrected code is
at \texttt{experiments/exp6\_measurement.py} and \texttt{experiments/exp6\_table3.py}.
\begin{table}[H]
\centering
\small
\caption{Action selection strategy comparison on sinusoidal signal (mean $\pm$ s.d.\ over
5 seeds). This table alone reports multiple seeds --- all three strategies were re-run
over 5 seeds for the correction; the forward-sampled baseline reproduces the preprint
result (seed-42: 80.7\%/$17.32\times$/0.788) within the distribution. The other tables
report the representative seed-42 run, matching the preprint.
$^{\dagger}$Spike for the locked-action baselines reflects a
\emph{constant-offset artifact}: $\mathrm{pred\_A}$ hardcodes the locked action value, so
replacing it with zero produces large error regardless of causal understanding. Spike is
therefore not diagnostic here; the verdict rests on pred gap and autocorr.}
\label{tab:strategy}
\begin{tabular}{@{}lrrrl@{}}
\toprule
Strategy & pred gap & Spike & Autocorr & Behavior \\
\midrule
Forward-sampled & $84.1{\pm}5.5\%$ & $16.6{\pm}2.9\times$ & $0.779{\pm}0.015$ & Structured \\
Direct AG gradient & $-4.9{\pm}8.5\%$ & $16.3{\pm}22.2\times^{\dagger}$ & $0.995{\pm}0.005$ & Locked \\
Gradient disagree & $0.4{\pm}0.6\%$ & $141.5{\pm}38.4\times^{\dagger}$ & $1.000{\pm}0.000$ & Locked \\
\bottomrule
\end{tabular}
\end{table}

\begin{figure}[htbp]
  \centering
  \includegraphics[width=0.9\linewidth]{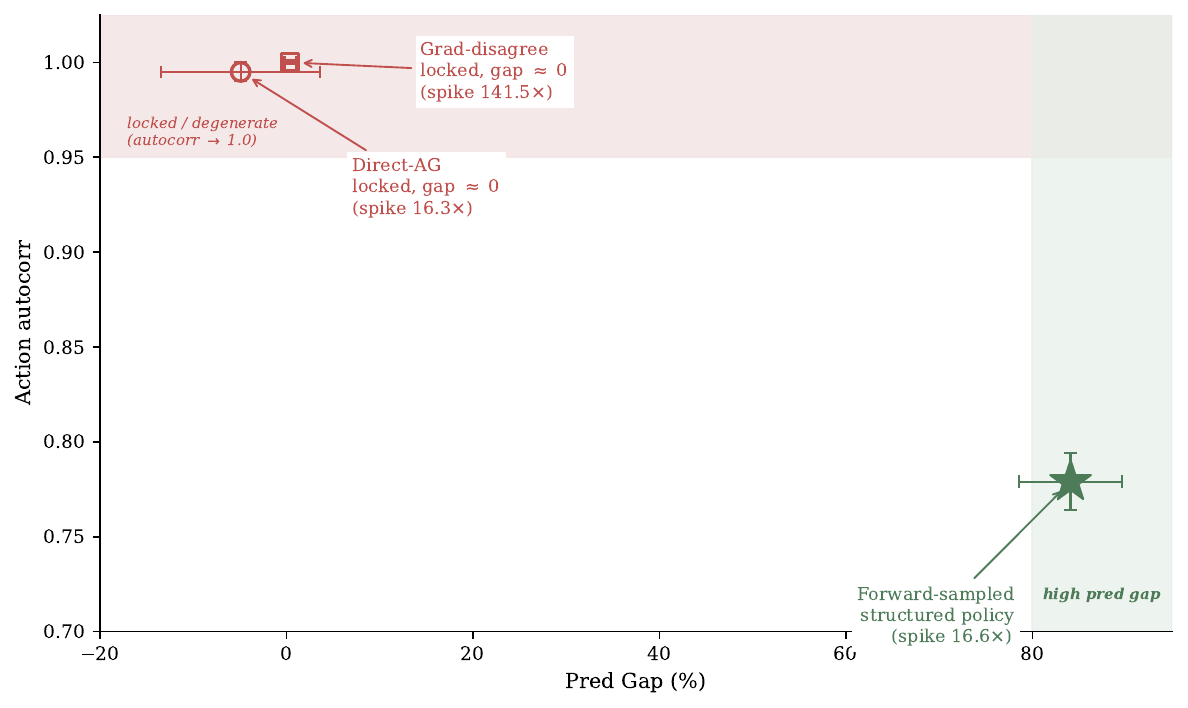}
  \caption{Action strategy comparison (mean $\pm$ s.d.\ over 5 seeds; bars are s.d.). Axes
    are the two diagnostic metrics: pred gap and action autocorr. Forward-sampled
    disagreement maximisation (star) reaches a high pred gap ($\sim$$84\%$) with a
    \emph{structured} policy (autocorr $\approx 0.78$); both gradient-based baselines
    collapse to a \emph{locked} action (autocorr $\approx 1.0$, red band) with pred gap near
    zero. The annotated spike ratios show why spike is \emph{not} diagnostic here --- the
    locked baselines carry the \emph{largest} spike ($\sim$$140\times$), a constant-offset
    artifact (Table~\ref{tab:strategy}).}
  \label{fig:strategy}
\end{figure}

Optimizing agency gain as a gradient objective on the policy collapses it to a
\emph{locked} action (Table~\ref{tab:strategy}: autocorr $\to 1.0$, pred gap $\to 0$):
with the action held nearly constant, there is no self-caused variation for the metric to
reward. More precisely, agency gain measures the predictive value of action
\emph{variation}, not action \emph{existence}: a constant action is absorbed into
\texttt{pred\_B}'s baseline, eliminating the advantage. The spike ratio is \emph{not} diagnostic here --- a locked constant offset inflates
it (up to ${\sim}140\times$) without any genuine causal structure --- so the verdict rests
on pred gap together with autocorr. Only forward-sampled disagreement maximisation yields a
\emph{structured} policy (autocorr $\approx 0.78$) with a genuine agency gain, robustly
across five seeds (Figure~\ref{fig:strategy}).

\textbf{Ablation: proprioception.}
\begin{table}[H]
\centering
\small
\caption{Agency gain with and without proprioceptive trace.}
\label{tab:ablation}
\begin{tabular}{@{}lrrl@{}}
\toprule
Config & pred gap & Spike & Autocorr \\
\midrule
With trace (obs+$\tau$, 5d) & 80.7\% & $17.32\times$ & 0.788 \\
Without trace (obs, 4d) & 81.1\% & $23.52\times$ & 0.766 \\
\bottomrule
\end{tabular}
\end{table}

Agency gain survives removal of the proprioceptive channel (Table~\ref{tab:ablation}). Proprioception is an
enhancer for explicit self-representation (Section~\ref{sec:proprioception}) but not
a requirement for agency gain measurement. The slightly \emph{higher} spike without the
trace reflects redundancy rather than an improvement: when the trace is present,
$h_{\text{multi}}$ already carries a partial copy of the action, so ablating
\texttt{pred\_A}'s direct action input hurts less; without the trace, that input is the
model's only source of action information, and removing it is correspondingly more
damaging.

\textbf{Cross-signal validation.}
\begin{table}[H]
\centering
\small
\caption{Agency gain across signal types.}
\label{tab:signals}
\begin{tabular}{@{}lrrr@{}}
\toprule
Signal & pred gap & Spike & Autocorr \\
\midrule
Sinusoidal & 80.7\% & $17.32\times$ & 0.788 \\
Lorenz & 99.5\% & $141.95\times$ & 0.762 \\
\bottomrule
\end{tabular}
\end{table}

The higher absolute numbers on Lorenz reflect the signal's smaller normalized amplitude
relative to the action range, not a stronger causal effect. Agency gain is computable
and positive on both periodic and chaotic signals (Table~\ref{tab:signals};
Figure~\ref{fig:agency}).

\FloatBarrier
\section{Conditions for Self-World Decomposition}

\subsection{Sufficient Conditions}

Table~\ref{tab:conditions} lists the four conditions, each with what fails in its absence.

\begin{table}[htbp]
\centering
\small
\caption{Four sufficient conditions for self-world decomposition, in developmental order.}
\label{tab:conditions}
\begin{tabular}{cl >{\raggedright\arraybackslash}p{3cm} >{\raggedright\arraybackslash}p{4cm}}
\toprule
\# & Condition & Without it & With it \\
\midrule
1 & Persistent state (never reset) & No attractors & Stable structure (6/7 scorecard) \\
2 & Causal action loop & No agency & Recovery 74.8\% vs.\ Control 57.2\% \\
3 & Proprioceptive feedback & Encoding gap (trailing 12.3\%) & Explicit representation (trailing 56.5\%) \\
4 & Asynchronous awakening & Fragile, LR-dependent & Robust: spike $5.58\times$, trailing 66.3\% \\
(5) & Dual-head (measurement) & No quantitative metric & pred gap 80.7--99.5\%, spike 17--142$\times$ \\
\bottomrule
\end{tabular}
\end{table}

Conditions 1--4 build on one another in order. Proprioception without a causal loop
provides no action information to encode. Asynchronous awakening without stable
perception provides a less robust foundation for action learning. The overall developmental
chain is illustrated in Figure~\ref{fig:chain}.

\subsection{Twelve Falsified Hypotheses}
\label{sec:falsified}

Each of the following hypotheses was a plausible candidate pathway;
all twelve were tested and falsified, delineating the boundary of what the
minimal system can and cannot achieve. Two of them (H10, H12) further probe whether the
account extends to a nonlinear 2-D spatial-action setting rather than the 1-D signal used
throughout; both fail.

\begin{table}[htbp]
\centering
\small
\caption{Twelve falsified hypotheses: what was tried, why it fails, and the evidence.}
\label{tab:falsified}
\begin{tabular}{c >{\raggedright\arraybackslash}p{3.2cm} >{\raggedright\arraybackslash}p{3.8cm} >{\raggedright\arraybackslash}p{3.2cm}}
\toprule
\# & Hypothesis & Why it fails & Evidence \\
\midrule
1 & FC network $\to$ modularity & No selection pressure & Self/world overlap $\approx$ random \\
2 & Stronger action $\to$ explicit encoding & Implicit compensation scales with action & $+1.9$pp only \\
3 & Passive EMA $\to$ post-action memory & Exponential decay, no reactivation & Ceiling 34\% \\
4 & Complex probe for weak signals & Collapses to majority class & GRU probe 1.4\% \\
5 & Character-level CE $\to$ rare token learning & Gradient dominated by 99\% majority & I-accuracy 0.0\% \\
6 & Trailing self-sustains without aux loss & Weak physical anchor & BA $80\% \to 70\%$ \\
7 & Single channel $\to$ other-agent model & Blind source separation at low SNR & B-trailing $<3\%$ \\
8 & Awareness + intention co-learnable & Non-stationary target (closed loop) & LR dilemma \\
9 & Gumbel-Softmax $\to$ discrete action & Jacobian disperses signal & Autocorr $= -0.007$ \\
10 & Gradient-trained policy $\to$ spatial action & Policy collapses to low-diversity actions & Spike $2.13 \to {<}1.6$ \\
11 & Residual perception $\to$ better attribution & Stillness-detector shortcut & Spike inverts \\
12 & Counterfactual + burst gate (2-D grid; stillness lacks action channel) & No valid action channel during quiet periods & Phase 2 collapses \\
\bottomrule
\end{tabular}
\end{table}

\subsubsection*{Detailed Analysis of Falsified Hypotheses}

Below we expand each hypothesis with the specific mechanism of failure, providing
context beyond the compressed table.

\paragraph{H1: Fully-connected network produces modularity.} A sufficiently large FC
layer over $h_{\text{multi}}$ was hypothesized to spontaneously develop separated
self/world subspaces. Result: self and world dimensions overlapped at chance level.
Without selection pressure toward separation, distributed representations remain
entangled.

\paragraph{H2: Stronger action produces explicit encoding.} Increasing $\gamma$ from
2.0 to 3.0 was hypothesized to strengthen the self-signal enough to encode explicitly.
Result: trailing recall improved by only 1.9pp. Implicit compensation scales with
action magnitude, but the encoding gap is architectural, not signal-strength-limited.

\paragraph{H3: Passive EMA sustains post-action memory.} Adding a slow-decay EMA
readout after the GRU was hypothesized to retain action history. Result: ceiling at
34\% recall. Exponential decay progressively erases the trace; without an active
reactivation mechanism, EMA cannot preserve the signal across trailing windows.

\paragraph{H4: Complex probes extract weak signals.} A nonlinear GRU-based probe was
hypothesized to extract information that linear probes cannot. Result: 1.4\% recall
(worse than linear). Complex probes overfit noise and collapse to predicting the
majority class; if the signal is not encoded, no probe complexity can extract it.

\paragraph{H5: Character-level cross-entropy grounds ``i'' tokens.} Character-level
prediction with rare ``i'' tokens was hypothesized to develop first-person grounding.
Result: 0.0\% i-accuracy. The gradient is dominated by the 99\% majority class of
world tokens; rare tokens receive no learning signal.

\paragraph{H6: Trailing signal self-sustains without auxiliary loss.} Once formed
with aux loss, the trailing encoding was hypothesized to persist under prediction
pressure alone. Result: balanced accuracy drops from 80\% to 70\% when aux loss is
removed. The trailing signal has weak physical anchor in the causal loop; without aux
pressure, it degrades (though not as catastrophically as the Control condition in
Section~\ref{sec:self-maintenance}).

\paragraph{H7: Single-channel setup enables other-agent modeling.} With sufficient
training, the system was hypothesized to distinguish ``another agent acted'' from
``I acted'' using a single observation channel. Result: B-trailing (other-agent
detection) below 3\%. Blind source separation at low SNR is not solvable by the GRU
architecture without additional information channels.

\paragraph{H8: Awareness and intention are co-learnable.} Training $W_{\text{action}}$
jointly with perception was hypothesized to yield robust results. Result: LR dilemma
--- no single learning rate is robustly best on both spike and trailing recall, and results swing with the rate (Table~\ref{tab:simul}).
The failure is \emph{not} gradient interference: perception and action are optimized by
separate optimizers over disjoint parameter sets, so no parameter is updated by both
objectives. The cause is \emph{non-stationarity of the learning target}. As
$W_{\text{action}}$ changes, the executed actions change, which shifts both the
observation distribution ($\text{obs}[0] \mathrel{+}= \gamma \cdot a$) and the
proprioceptive trace, and therefore the trajectory of $h_{\text{multi}}$ itself; the
perceptual readouts are chasing a moving target. The coupling is bidirectional, since
action selection depends on the very heads being trained. A fast action learning rate
destabilizes perception; a slow one yields a weak policy --- hence the dilemma.
Temporal separation resolves it by letting perception consolidate against a
\emph{stationary} action distribution (Phase~1 random actions) before the policy is
allowed to move.

Hypotheses H9--H12 concern implementation-specific alternatives (Gumbel-Softmax discrete
actions, a gradient-trained spatial policy, residual perception, and counterfactual
training on a 2-D grid). Their detailed failure analyses are collected in
Appendix~\ref{app:extended-h}.

\medskip

Together, these 12 hypotheses map the boundary of what the minimal system architecture
supports --- and, equally important, what it does not. Each represents 1--2 weeks of
exploratory work; making these negative results explicit saves future work from
repeating them.

\section{Discussion}

\subsection{Unified Information-Theoretic Framework}
\label{sec:discussion-unified}

The six developmental stages can be unified under an information-theoretic
interpretation. Each stage corresponds to the emergence of a distinct
information-theoretic quantity. In the notation below, $H_t$ denotes the complete
persistent state $h_{\text{multi}}$, $Z_t = g(H_t)$ is its compressed representation
extracted by the readout heads, $O_{t+1}$ is the next observation, and $A_t$ the
system's action; $S_t^{\text{self}}$ denotes the system's self-action state (currently
or recently acting).

\textbf{Stage 1 --- Perception:} The system learns a compressed latent $Z_t = g(H_t)$
that maximizes $I(Z_t;\, O_{t+1})$ (attractor formation, 6/7 scorecard).

\textbf{Stage 2 --- Causation:} Action creates conditional mutual
information $I(O_{t+1}; A_t \mid H_t) > 0$
(channel-specific spike, confirmed causal).

\textbf{Stage 3 --- Encoding gap:} The causal information exists in the system (Stage 2)
but is not encoded in the state: $I(Z_t;\, S_t^{\text{self}}) \approx 0$, where
$S_t^{\text{self}}$ denotes the system's self-action state (whether it is currently
or recently acting). The system
compensates for its actions through distributed weight dynamics without creating an
explicit self-variable (trailing recall 12.3\%, near chance; the overall
action-state probe reaches 70\% only because the easy active and quiet states
inflate it).

\textbf{Stage 4 --- Awareness:} Proprioceptive feedback makes the implicit explicit:
$I(Z_t;\, S_t^{\text{self}}) \gg 0$ (trailing recall $12.3\% \to 56.5\%$, symbol
grounding BA $80.1\%$).

\textbf{Stage 5 --- Intention:} Under Gaussian assumptions, agency gain approximates
the conditional mutual information:
\begin{equation}
  \mathcal{A} \approx I(O_{t+1};\, A_t \mid H_t)
    = H(O_{t+1} \mid H_t) - H(O_{t+1} \mid H_t, A_t)
    \approx \text{Err}_{\text{world}} - \text{Err}_{\text{self}}
\end{equation}

\textbf{Stage 6 --- Measurement:} The spike test implements a do-operation:
$\text{do}(A_t = 0)$. A spike confirms that the measured agency gain reflects genuine
causal influence, not statistical correlation~\cite{pearl2009}.

Stages 2--5 can be read as successive attempts to close a single dissociation: the
system's use of its own action in prediction ($I(O_{t+1}; A_t \mid H_t)$, positive from
Stage~2) runs ahead of its representation of that action in the readable state
($I(Z_t; S_t^{\text{self}})$, near zero until Stage~4). The encoding gap is the point
where this dissociation is most visible; the four sufficient conditions are the steps
that close it.

Table~\ref{tab:compare} contrasts agency gain with related measures: unlike empowerment
and free energy it needs no density estimation, and unlike all three it is explicitly
self/world-decomposing.

\begin{table}[h]
\centering
\small
\caption{Comparison of agency measures.}
\label{tab:compare}
\begin{tabular}{@{}lccc@{}}
\toprule
Measure & Density est. & Diff. & Self/world \\
\midrule
Empowerment~\cite{klyubin2005} & Required & Approx. & No \\
Curiosity~\cite{pathak2017} & No & Yes & No \\
Free energy~\cite{friston2010} & Required & Approx. & Implicit \\
\textbf{Agency gain} & \textbf{No} & \textbf{Yes} & \textbf{Yes} \\
\bottomrule
\end{tabular}
\end{table}

\subsection{When the GRU Is a Fixed Reservoir}
\label{sec:discussion-reservoir}

Whether the GRU is trained depends on the configuration, and the distinction matters
for how the results should be read. The GRU is a fixed reservoir in the perception,
causal-budding, and measurement experiments (Sections~\ref{sec:perception},
\ref{sec:causal}, and~\ref{sec:measurement}, excluding the self-maintenance ablation of
Section~\ref{sec:self-maintenance}): its weights are randomly initialized and receive
zero gradient through the detach mechanism (Section~\ref{sec:training}); only the
readout heads are trained; $W_{\text{action}}$ is a fixed random policy in perception and
causal budding, and is trained only in the measurement configuration.
In every experiment that instead applies an auxiliary classification loss to the
\emph{live} state --- the self-maintenance ablation
(Section~\ref{sec:self-maintenance}) and
Sections~\ref{sec:encoding}--\ref{sec:async} --- gradient \emph{does} propagate into
the GRU during Phase~2.

In the measurement configuration, therefore, the information in $h_{\text{multi}}$ is
not learned --- it is a natural consequence of the reservoir's fixed dynamics
propagating input structure into its state space. The readout heads discover and
extract this information, but do not create it. This strengthens the claim that the
dual-head architecture is a measurement instrument rather than an information source:
agency gain reflects a real property of the reservoir-environment interaction, not an
artifact of training.

The same distinction makes the encoding-gap result \emph{stronger} than a readout
limitation. In Section~\ref{sec:encoding} the auxiliary loss actively propagates
gradient into the recurrent weights, explicitly pressuring the GRU to encode
``recently acted'' --- and trailing recall still remains at 12.3\%. Insufficient
training cannot explain this: Section~\ref{sec:proprioception} applies the
\emph{identical} auxiliary loss to the \emph{identical} target under the
\emph{identical} training budget, differing only by a single additional input
dimension, and reaches 56.5\%. The gap therefore reflects neither a fixed reservoir
failing to carry a signal, nor a probe too weak to extract it, nor an optimizer given
too little pressure: even when the recurrent dynamics themselves are trained toward
the target, the representation cannot form without a dedicated input channel. Only architectural
intervention (adding the proprioceptive trace channel,
Section~\ref{sec:proprioception}) closes it.

\subsection{The Dual-Head Architecture: Thermometer, Not Source}
\label{sec:discussion-thermometer}

A natural objection is that the dual-head design creates rather than detects self-world
decomposition. This objection conflates opportunity with realization. The architecture
guarantees only that \texttt{pred\_A} has access to action information. It does not
guarantee that \texttt{pred\_A} develops any genuine causal dependence on that action:
the policy may stay unstructured, and nothing forces the system to choose actions that
keep its causal role identifiable. These are emergent properties of
training --- as the Phase~1 baseline below shows, action access alone produces a
large gap through mechanical compensation, but no causal dependence.

The Phase 1 baseline directly addresses the ``unfair prior'' objection.
Even under random actions, \texttt{pred\_A} achieves a large prediction gap ---
but this reflects mere mechanical compensation of a known perturbation, not causal
agency. The two phases are distinguished not by gap magnitude but by the spike ratio
(the error increase when the action input is removed):

\begin{center}
\small
\begin{tabular}{@{}lrr@{}}
\toprule
& Pred gap & Spike \\
\midrule
Phase 1 (random actions)  & 98.8\% & $0.95\times$ \\
Phase 2b (trained policy) & 80.7\% & $17.32\times$ \\
\bottomrule
\end{tabular}
\end{center}

The gap fails to distinguish the two phases --- both are 80--99\%.
The spike does: a Phase~1 spike of $0.95\times$ (indistinguishable from~1.0)
shows the system has not yet formed a functioning causal loop, while the
Phase~2b spike of $17.32\times$ confirms the trained policy generates actions
\texttt{pred\_A} genuinely depends on. Agency gain requires not just access to
action information, but that the information be causally structured.

The low Phase~1 spike does not mean \texttt{pred\_A} lacks action dependence
--- it has learned $w_a \approx \gamma$ from the large random actions used
during training. Rather, it reflects the fact that the untrained
$W_{\text{action}}$ produces near-zero actions during the spike test, so the
dependence exists but is undetectable at that scale. What distinguishes
Phase~2b is not that \texttt{pred\_A} has learned to use actions (it already
has), but that $W_{\text{action}}$ has developed a policy that produces
actions large and structured enough for the dependence to matter.

Formally, the spike test is an \textbf{interventional (Pearl Layer~2) probe}: it
intervenes on \texttt{pred\_A}'s action input --- replacing the executed action $a$ with
$0$ (or $-a$) while the observed sequence is held fixed --- and compares
$\mathbb{E}[\text{Err}_A \mid \text{do}(a_{\text{in}}{=}0)]$ against
$\mathbb{E}[\text{Err}_A]$. Because it manipulates the input rather than merely observing
it, a large spike ($17.32\times$ on sinusoidal, $141.95\times$ on Lorenz), read against
the Phase~1 baseline where the identical intervention leaves the error unchanged
($0.95\times$), establishes that the trained predictor's output is \emph{causally} (not merely correlationally) dependent on the self-generated action, and that this
dependence is a trained property rather than an artifact of action access.
This comparison establishes that predictive advantage and causal dependence are
separable: Phase~1 achieves the former without the latter.

\textbf{Relation to forward-model theory.} The dual-head configuration is structurally
close to the classical \emph{forward model} of motor control~\cite{vonholst1950,
wolpert1998, wolpert2000}: \texttt{pred\_A}$(h, a)$ receives an action copy and predicts
its sensory consequences (as a forward model does using efference copy), while
\texttt{pred\_B}$(h)$ serves as a counterfactual baseline lacking that copy. The same
predict-then-compare motif also underlies predictive-coding accounts of cortical
processing~\cite{rao1999, friston2010}. The novelty of this work is not the
architectural motif itself; it is the \emph{emergence question}: rather than positing a
forward model and training it with an explicit (state,~action)~$\to$~next-state
objective, we ask under what minimal conditions a system trained solely on
prediction error develops forward-model-like structure as a computational property of
its state dynamics, with agency gain acting as the read-out through which the property
becomes measurable.

\subsection{The Encoding Gap: Prediction \texorpdfstring{$\neq$}{!=} Self-Representation}

The encoding gap (Section~\ref{sec:encoding}) is perhaps the most important finding for
theories of self-awareness. A system can achieve strong implicit agency — perfectly
compensating for its own actions in prediction — while its readable self-state is nearly
absent: trailing recall (detecting ``recently acted'' after the action stops) sits at
12.3\%, near chance. The overall self-classification probe reaches 70\% only because the
trivially separable active and quiet states inflate it. Six experiments confirm this is not a measurement
limitation but a genuine representational absence.

This dissociation is not merely an empirical observation but a specific prediction:
self-representation requires both \emph{informational value} (present through the causal
loop) AND a dedicated \emph{architectural pathway} through which that information can
enter the system's state (missing without the proprioceptive trace channel). The
encoding gap is precisely what the absence of the second condition looks like when the
first is already satisfied.

The same dissociation should recur wherever a system compensates for its own actions
without a dedicated channel writing them back into its state: competent control does
not, by itself, produce an explicit self-model, and the two capacities can succeed or
fail independently.

\subsection{Asynchronous Awakening: A Robust Strategy}

The finding that temporal separation produces the most robust results reflects a
practical constraint on shared-representation systems. When perception and action train
simultaneously, results are fragile and sensitive to learning rate. Asynchronous
training consistently produces the highest spike and trailing recall across all
configurations.

This parallels observations in infant development: stable sensory processing precedes
intentional motor control by months~\cite{rochat2001}. Our minimal system suggests this
developmental sequence may reflect an optimization landscape constraint.

\subsection{Structural Correspondence with the Yog\={a}c\={a}ra Layered-Consciousness Model}

While all architectural decisions were made on engineering grounds, the emergent
developmental sequence exhibits a structural parallel with the layered-consciousness
model of the Yog\={a}c\={a}ra (Weishi, Consciousness-Only) school of
Buddhism~\cite{lusthaus2002}. We note these as parallels in computational function, not
in experience; we make no claims about consciousness or phenomenal selfhood.

The system begins with no world model and no self-representation; everything it comes to
encode arises from the stream of external stimulation it receives --- a starting point
that resonates with the Yog\={a}c\={a}ra view that cognition is constituted through the
contact of the sense faculties with their objects, not given in advance. From this
starting point, the developmental sequence unfolds as follows.

The never-resetting $h_{\text{multi}}$, which carries and accumulates traces across all
steps but does not itself perform the self/world discrimination --- that arises from the
contrast between the readout heads --- parallels the \={a}laya-vij\~n\={a}na (storehouse
consciousness): a substrate of moment-to-moment continuity that holds what has passed
but does not itself judge. The proprioceptive channel, through which implicit causal
knowledge becomes explicit (Section~\ref{sec:proprioception}), parallels
k\={a}ya-indriya (the body faculty) --- the sense through which the system comes to
register its own activity. The self-appropriating function of manas --- continuously
grasping the substrate's content as ``mine'' --- parallels the learnable action policy
$W_{\text{action}}$, which reads the persistent state but cannot alter it (the detach
boundary ensures that action-learning gradients do not flow back into the perceptual
substrate). The emergent behavior of selecting actions to keep self-caused events
predictable (not prescribed by the architecture, but arising from training) is the
functional expression of that grasping.

These three are correspondences recognized after the experiments, not design principles
imported before them. One element of the sequence is different in kind: the strict
ordering of stable perception before intentional action was not a post hoc observation
but a prospective constraint --- three simultaneous-learning configurations produced
fragile results sensitive to learning rate (Section~\ref{sec:async}); only temporal
separation yielded consistently high performance across both metrics. That this
independently-derived ordering coincides with the contemplative principle of
\'samatha (concentration) before vipa\'syan\={a} (insight) is, to us, the
correspondence with the strongest empirical support --- though not a strict necessity,
since simultaneous learning can succeed under favorable conditions. We leave open whether these
parallels point to a wider constraint on self-representation or are peculiar to this
architecture.

\subsection{Self-Representation Is Sustained Only When Causally Useful}
\label{sec:discussion-sustained}

The self-maintenance result (Section~\ref{sec:self-maintenance}) reveals a selection pressure on
representations. When the encoding of burst-active state serves ongoing prediction
through the causal loop, the GRU's learned dynamics maintain it without any external
training signal: removing the auxiliary loss leaves accuracy at 94.9\%.
When the same encoding has no predictive utility (as in the control group, where the action statistics are matched but the causal loop is absent), it decays the moment
the training signal is removed, collapsing to 53.9\%.
This asymmetry suggests that causal grounding is not merely a correlate of
self-representation but a precondition for its persistence.

A subsequent spiking-neuron study~\cite{han2026} builds on this framework, reimplementing
the agency comparator on a Nengo LIF/PES substrate (indicating that the comparator is not
specific to the recurrent architecture used here) and extending the account toward
\emph{behavioral} durability through agency-gated credit assignment. This is complementary to the
\emph{representational} durability established above: the self-encoding persists without
supervision precisely when it remains causally useful.

\section{Limitations}
\label{sec:limitations}

The system operates on 4-channel sinusoidal signals and Lorenz chaotic attractors with
fewer than 100K parameters. Actions take effect immediately ($\tau = 0$) in most
configurations; the self-maintenance experiment (Section~\ref{sec:self-maintenance}) is
the exception, using a two-step delay. In the measurement configuration, the aligned
action value is passed directly to \texttt{pred\_A}, so agency gain does not test
temporal causal reasoning per se. Genuine temporal delay robustness --- requiring the
system to attribute observations to actions across time lags rather than receive the
aligned action explicitly --- remains future work.

\textbf{Scale dependence.} The absolute magnitudes of prediction gap and spike ratio
depend on the ratio of action range to signal amplitude. With action range $= 2.0$ and
signal amplitudes of order 1, the action dominates observable variance, producing large
gap values. The scientifically meaningful comparisons are relative: Causal vs.\ Control,
with-trace vs.\ without-trace, forward sampling vs.\ gradient methods, async
vs.\ simultaneous. These relative differences are invariant to scaling.

\textbf{Reservoir dimensionality.} The hidden dimension of 192 is a pragmatic
engineering choice for the signals studied here (4-channel sinusoids and Lorenz),
where the effective attractor dimensionality is small (5 out of 192,
Section~\ref{sec:perception}). In principle, reservoir capacity should scale with
the intrinsic complexity of the target signal --- a well-established relationship in
reservoir computing~\cite{jaeger2001}, where memory and separation properties are
bounded by network size. Systems facing higher-dimensional or longer-timescale inputs
would require correspondingly larger reservoirs, echoing the observation that
biological neural systems scale their internal state dimensionality with the
complexity of their sensorimotor demands. The developmental conditions identified
here are architectural in nature and should generalize across reservoir sizes; the
specific dimension chosen is not.

\textbf{Action autocorrelation.} The temporal coherence of Phase 2b actions
(autocorrelation $\sim$0.8) arises partly from $W_{\text{action}}$ reading the
slowly-varying $h_{\text{multi}}$, and partly from training. The relative contribution
of training vs.\ structural correlation is not fully disentangled.

Self/other discrimination fails: the system can distinguish ``I caused this'' from
``something else caused this'' but cannot identify which other agent caused a change.
Two-dimensional intentional action fails across five versions due to gradient conflict
in spatial action learning.

Agency gain \emph{as a metric} measures predictive advantage under observational
distributions; the interventional evidence for causal dependence is provided separately
by the spike test (Section~\ref{sec:discussion-thermometer}), a Pearl Layer~2 probe on
the action input~\cite{pearl2009}. A fully causal formulation of the metric itself
remains future work.

We make no claims about consciousness, sentience, or subjective experience.

\section{Conclusion}

We have traced the development of self-world decomposition in a minimal predictive
system through six developmental stages and twelve falsified alternatives. The
developmental path --- from stable perception, through implicit causal attribution,
past an encoding gap, to explicit self-representation and finally quantifiable agency
gain --- is ordered: each stage builds on the conditions established by the previous, with
asynchronous awakening the most robust route rather than the only one.

Four conditions are sufficient, in order: persistent state, causal action loop,
proprioceptive feedback, and asynchronous awakening. Twelve alternative approaches
fail, for reasons we characterize precisely. The core results are relative comparisons
that do not depend on scaling: Causal recovery $74.8\%$ vs.\ Control $57.2\%$
(Section~\ref{sec:causal}), trailing recall $12.3\% \to 56.5\%$ with proprioception
(Section~\ref{sec:proprioception}), async spike $5.58\times$ vs.\ simultaneous best $4.76\times$ (Section~\ref{sec:async}), forward-sampled agency gain positive while
two gradient alternatives degenerate (Section~\ref{sec:measurement}), and
self-representation is sustained only when causally useful (Causal $94.9\%$
vs.\ Control $53.9\%$ after the training signal is removed,
Section~\ref{sec:self-maintenance}).
Notably, all core results replicate on both regular (sinusoidal) and chaotic (Lorenz)
signals, ruling out that our findings depend on specific spectral properties of the
environment.

The 12 falsified hypotheses are arguably the most transferable contribution of this
work. They map the boundary between systems that predict and systems that know they
are the ones predicting, and each represents 1--2 weeks of exploratory work whose
failure mechanism is now precisely characterized --- saving future work from repeating
these attempts without first understanding why they failed in this minimal system.

Agency gain — the predictive advantage of knowing one's own action — is the metric that
makes this developmental process quantifiable. Its absolute magnitude depends on
action-to-signal scaling (see Section~\ref{sec:limitations}), but its sign, its sensitivity to ablation,
and its relative behavior across conditions constitute the empirical evidence of this
paper. A methodological corollary generalizes beyond our setting: agency-related metrics
constructed via loss differencing are subject to Goodhart-type gaming when used as
gradient objectives (Section~\ref{sec:measurement}, action-strategy comparison), and
require selection mechanisms external to the gradient (forward sampling in our case) to
avoid degeneracy.

Different architectures and richer environments will show whether the constraints
identified here — the encoding gap, the proprioceptive breakthrough, the ordering of
awareness before intention — are universal principles of self-representation or
particular to the GRU.

\appendix

\section{Extended Analysis of Implementation-Specific Hypotheses}
\label{app:extended-h}

The four hypotheses below concern specific implementation alternatives that were tested
and falsified. Their compressed summaries appear in Table~\ref{tab:falsified}; the
detailed failure mechanisms are given here.

\paragraph{H9: Gumbel-Softmax enables discrete action learning.} Replacing continuous
action with Gumbel-Softmax was hypothesized to produce structured discrete behavior.
Result: action autocorrelation $= -0.007$ (essentially random). The Jacobian of the
softmax disperses the training signal across action categories, preventing coherent
policy formation.

\paragraph{H10: A gradient-trained policy learns nonlinear spatial action.} On the 2-D
grid the action-to-observation mapping is nonlinear (a move shifts the entire
dense-texture view), unlike the linear $\text{obs}[0]\mathrel{+}=\gamma a$ of the 1-D
signal; with random actions, causal attribution already succeeds here (spike $2.13$). We
hypothesized that letting the policy $W_{\text{action}}$ learn from the prediction
gradient --- an action-conditioned prediction head whose error also trains the policy ---
would extend structured spatial action to this setting. Result: spike drops to $<1.6$. The
prediction objective favors \emph{predictable} actions, so the gradient collapses the
policy toward low-diversity behavior (near-constant direction; action s.d.\ $\approx0.04$),
and forcing diversity with an activity constraint trades directly against the spike
($1.52 \to 1.39$). This is the same policy collapse that gradient-based action objectives
produce in the 1-D measurement setting (Section~\ref{sec:measurement}), and it is resolved
the same way --- by \emph{forward-sampling} candidate actions rather than training the
policy through the prediction gradient.

\paragraph{H11: Residual perception improves attribution.} Predicting a residual
after world model subtraction was hypothesized to sharpen self-attribution. Result:
spike ratio inverts (indicating opposite effect). The residual becomes a
stillness-detector shortcut; the model learns to predict ``no change'' rather than
``self-caused change''.

\paragraph{H12: Counterfactual training with burst gate on 2-D grid.} Pairing burst
gating with counterfactual directional prediction on a 2-D grid was hypothesized to
produce explicit self-modeling. Result: Phase~2 collapses (no learning). During quiet
periods, the 2-D grid lacks a valid action channel, so counterfactual comparison
becomes ill-defined and gradients destabilize.

\section{Perception Scorecard}
\label{app:scorecard}

The perception experiment (Section~\ref{sec:perception}) is evaluated against seven
pre-specified tests of stable attractor formation (Table~\ref{tab:scorecard}). Six pass;
the exception is the residual-whiteness test, indicating the GRU does not fully extract
all predictable structure from the signal. The other experiments state their PASS/FAIL
thresholds inline (e.g.\ trailing recall, spike ratio) rather than as an aggregate
scorecard.

\begin{table}[H]
\centering
\small
\caption{Perception scorecard: seven pre-specified tests of attractor formation
(Section~\ref{sec:perception}).}
\label{tab:scorecard}
\begin{tabular}{c >{\raggedright\arraybackslash}p{2.5cm} >{\raggedright\arraybackslash}p{4.8cm} l}
\toprule
\# & Test & Pass criterion & Result \\
\midrule
1 & Low dimensionality & effective dimension $<30\%$ of hidden dimension & PASS ($5/192$) \\
2 & Power-law inertia & PC1 autocorrelation fits a power law better than an exponential & PASS \\
3 & Perturbation recovery & recovery $>50\%$ after noise injection & PASS ($95.0\%$) \\
4 & Residual whiteness & prediction residual is white (no autocorrelation) & \textbf{FAIL} \\
5 & Novelty response & $h_{\text{multi}}$ displacement $>1.0$ on a frequency change & PASS \\
6 & Spectral separation & the four EMA groups have distinct peak frequencies & PASS \\
7 & Error stationarity & prediction error is stationary over time & PASS \\
\bottomrule
\end{tabular}
\end{table}

\section{Notation and Terminology}
\label{app:notation}

\subsection*{Notation}

\begin{table}[H]
\centering
\small
\begin{tabular}{ll}
\toprule
Symbol & Meaning \\
\midrule
$h_{\text{multi}}$ & Multi-scale EMA persistent hidden state (never reset) \\
\texttt{pred\_A} / \texttt{pred\_B} & Action-conditioned prediction head / action-blind prediction head \\
$A$ (agency gain) & $\text{Err}_{\text{world}} - \text{Err}_{\text{self}}$ \\
pred gap & $(\text{Err}_{\text{world}} - \text{Err}_{\text{self}}) / \text{Err}_{\text{world}}$ \\
spike & $\text{Err}_{\text{self}}(\text{action disconnected}) / \text{Err}_{\text{self}}(\text{normal})$ \\
$\tau$ & Action trace: $\tau(t) = 0.95\cdot\tau(t{-}1) + 0.05\cdot|a(t)|$ \\
$\alpha$ & EMA timescale coefficient, ranging $0.02$–$0.80$ \\
$\gamma$ & Action gain on $\text{obs}[0]$, fixed at $2.0$ \\
$W_{\text{action}}$ & Linear projection from $h_{\text{multi}}$ to action \\
\bottomrule
\end{tabular}
\end{table}

\subsection*{Terminology}

\begin{table}[H]
\centering
\small
\begin{tabular}{>{\raggedright\arraybackslash}p{4.8cm}
                >{\raggedright\arraybackslash}p{9.5cm}}
\toprule
Term & Definition \\
\midrule
Self-world decomposition &
  Distinguishing observation changes caused by the system's own actions from
  those caused by the external world. \\
Self-representation &
  A state variable in $h_{\text{multi}}$ that is \emph{causal} (its presence improves
  prediction of self-caused observations), \emph{readable} (linearly decodable from the
  persistent state), and \emph{persistent} (maintained by the system's own dynamics
  without ongoing supervision). \\
Agency gain &
  The predictive advantage that comes from knowing one's own action:
  how much better \texttt{pred\_A} predicts than \texttt{pred\_B}. \\
Agency &
  A system's capacity to causally influence its observations through its actions;
  operationally, both agency gain $A > 0$ \emph{and} a spike significantly above~$1.0$.
  Agency gain is the metric, the spike test the causal verification, agency the
  property they jointly diagnose. \\
Spike test &
  An interventional check computed over a measurement window (3{,}000 steps): the
  action input to \texttt{pred\_A} is replaced by zero at each step and the mean error
  increase across the window is measured. A spike $\gg 1$ confirms causal dependence
  rather than a statistical artifact. \\
Encoding gap &
  The dissociation between implicitly compensating for one's actions in
  prediction and explicitly encoding ``I am acting'' as a readable state
  variable in $h_{\text{multi}}$. \\
Trailing recall &
  A probe's ability to detect ``recently acted'' after action ceases,
  during the 50-step window following burst-gate deactivation. \\
Burst gate &
  The mechanism alternating between active (action on) and quiet
  (action off) periods. \\
Trailing period &
  The 50-step window immediately after the burst gate closes. \\
Async awakening &
  Phased training in which perception consolidates first (Phases 1--2a)
  before action learning begins (Phase 2b). \\
Self-maintenance &
  The phenomenon whereby self-representation is spontaneously sustained
  by the system when — and only when — it is causally useful for
  prediction. \\
Forward-sampled disagreement maximisation &
  Action selection by evaluating four candidates and executing the one
  that produces the largest prediction gap between \texttt{pred\_A} and
  \texttt{pred\_B}. \\
\bottomrule
\end{tabular}
\end{table}

\section*{Acknowledgments}
The author used AI for language editing, manuscript preparation, and coding
assistance. AI tools were not used to generate the scientific content, data,
analyses, or conclusions of this study; the author takes full responsibility for the
integrity and accuracy of the work.

\section*{Funding}
This research received no external funding.

\section*{Author Contributions}
E.Y. conceived the study, designed and performed the experiments, analyzed the
results, and wrote the manuscript.

\section*{Competing Interests}
The author declares no competing interests.

\section*{Data Availability}
This study introduces no external empirical dataset. All signals analyzed are synthetic
(sinusoidal and Lorenz) and are generated deterministically by the code listed under
Code Availability, which reproduces every result and figure.

\section*{Code Availability}
The code to reproduce all experiments and figures in this version is available at
\url{https://github.com/yefeifei-tech/prediction-to-self-journal}. It is identical to the
arXiv~v1 code (\url{https://github.com/yefeifei-tech/prediction-to-self}) except for one
correction: the action-strategy comparison of Section~\ref{sec:measurement}
(Table~\ref{tab:strategy}, Figure~\ref{fig:strategy}), whose arXiv~v1 gradient-strategy
numbers were confounded. All other experiments, numbers, and figures are unchanged from
the preprint.

\printbibliography

\end{document}